\documentclass[letterpaper, 10 pt, conference]{ieeeconf}  % Comment this line out if you need a4paper

\IEEEoverridecommandlockouts                              % This command is only needed if 
\usepackage{cite}
\usepackage{subcaption}
\usepackage{amsmath,amssymb,amsfonts}
\usepackage{algorithmic}
\usepackage{graphicx}
\usepackage{textcomp}
\usepackage{xcolor}
\usepackage{booktabs} % For prettier tables
\usepackage{graphicx} % Required for including images
\usepackage{algorithmic}
\usepackage{algorithm}
\usepackage{array}
\usepackage{textcomp}
\usepackage{stfloats}
\usepackage{url}
\usepackage{verbatim}
\usepackage{longtable}
\usepackage{multirow}
\usepackage{graphicx}
\usepackage{cite}
\usepackage{hyperref}
\usepackage{parskip}
\usepackage{caption}

\makeatletter
\newcommand*\titleheader[1]{\gdef\@titleheader{#1}}
\AtBeginDocument{%
  \let\st@red@title\@title
  \def\@title{%
    \bgroup\normalfont\small\flushleft\@titleheader\par\egroup
    \vskip.5em\st@red@title}
}
\makeatother

\newcommand{\inet}{IncidentNet}

\newcolumntype{C}[1]{>{\centering\let\newline\\\arraybackslash\hspace{0pt}}m{#1}}

\title{\LARGE \bf
\inet{}: Traffic Incident Detection, Localization and Severity Estimation with Sparse Sensing
}

\author{Sai Shashank Peddiraju$^{1,*}$, Kaustubh Harapanahalli$^{2,*}$, Edward Andert$^{2}$ and Aviral Shrivastava$^{2}$\\% <-this % stops a 
Arizona State University, Tempe\\
\tt \small \{speddira,kharapan,eandert,aviral.shrivastava\}@asu.edu
\thanks{This work was partially supported by funding from National Science Foundation grants CPS 1645578 and Semiconductor Research Corporation (SRC) project 3154.}% <-this % stops a space
\thanks{${*}$: Equal Contributions from the authors}
\thanks{${1}$: School for Engineering of Matter, Transport and Energy, ${2}$: School of Computing and Augmented Intelligence}
}

\begin{document}

\titleheader{2024 IEEE 27th International Conference on Intelligent Transportation Systems (ITSC)}
\maketitle
\thispagestyle{empty}
\pagestyle{empty}

%%%%%%%%%%%%%%%%%%%%%%%%%%%%%%%%%%%%%%%%%%%%%%%%%%%%%%%%%%%%%%%%%%%%%%%%%%%%%%%%
\begin{abstract}

Prior art in traffic incident detection relies on high sensor coverage and is primarily based on decision-tree and random forest models that have limited representation capacity and, as a result, cannot detect incidents with high accuracy. This paper presents \inet{} - a novel approach for classifying, localizing, and estimating the severity of traffic incidents using deep learning models trained on data captured from sparsely placed sensors in urban environments. Our model works on microscopic traffic data that can be collected using cameras installed at traffic intersections. Due to the unavailability of datasets that provide microscopic traffic details and traffic incident details simultaneously, we also present a methodology to generate a synthetic microscopic traffic dataset that matches given macroscopic traffic data. 
\inet{}\footnote{\url{https://github.com/MPSLab-ASU/IncidentNet}} achieves a traffic incident detection rate of 98\%, with false alarm rates of less than 7\% in 197 seconds on average in urban environments with cameras on less than 20\% of the traffic intersections.
\end{abstract}

%%%%%%%%%%%%%%%%%%%%%%%%%%%%%%%%%%%%%%%%%%%%%%%%%%%%%%%%%%%%%%%%%%%%%%%%%%%%%%%%
\section{Introduction}

In 2019, traffic accidents alone caused approximately 28 million incidents, risking people's safety\cite{blincoe2022economic}. According to the study \cite{emergency_response} conducted across 2268 US counties, a 5-minute delay in emergency response increased fatality rates by 46\%, while response times under 7 minutes reduced fatality rates by 58\% in urban and rural areas. Along with traffic accidents, cargo spills, stalled vehicles, road maintenance, and other emergency scenarios are also traffic incidents. Traffic incidents are generally defined as non-recurring events that reduce the roadway's capacity\cite{pb_farradyne_inc_traffic_2000}. These incidents lead to secondary issues such as road congestion and delayed emergency support\cite{incident_stat}. This motivates the need to work towards detecting traffic incidents quickly, improving emergency response time, and improving traffic re-routing time.

Faster and more accurate incident detection presents two main challenges. (i) Need for an algorithm to detect, locate, and estimate the severity of incidents in urban regions: Most existing traffic incident detection algorithms, such as \cite{liang_traffic_2022}, are tailored for highways. However, the existing algorithms for urban regions, like \cite{yu_arterial_2015}, introduced an algorithm that compares current traffic conditions, including travel times, to a predefined threshold, and \cite{han_traffic_2020} proposed a pattern-matching algorithm that uses a database of GPS trajectories to identify incidents. However, the performance of such comparative and pattern-matching algorithms heavily depends on thresholds, requiring continuous adjustment due to traffic's dynamic nature. (ii) Non-availability of microscopic datasets: Existing well-known public datasets like PEMS\cite{PeMSwebsite}, San Francisco I-880\cite{skabardonis1996880}, and METR-LA\cite{metr_la} primarily use inductive loop detectors to capture macroscopic data focusing on highways by aggregating metrics like average vehicle speed and average flow-rate density obtained through these sensors without any vehicle distinguishing features. This level of aggregation makes it challenging to get high accuracy in dynamic urban settings. Alternatively, datasets using GPS sensors like NYC Taxi Data\cite{ren_graph_2024} and Bluetooth sensors like Highway 99-W\cite{yu_arterial_2015} offer microscopic features but suffer from issues like signal loss and interference and data latency\cite{GPS_study_traffic,bluetooth_devices_study}, hindering their use in time-critical traffic incident detection tasks.

Owing to the vast development of Computer Vision and the quality of cameras over the last decade, the deployment and utility of cameras for traffic use cases have increased in urban environments and highways\cite{advantage_of_cv}. They can capture microscopic data like speed, location, timestamp, direction, and unique vehicle identifiers for each vehicle. Due to this, developments have focused on traffic incident detection approaches within the camera's field of view\cite{shah2018accident}. However, incidents outside their field of view remain undetected. Deploying cameras to increase coverage to 100\% is challenging and not desirable. So, in this paper, we develop methods to identify incidents outside the camera's field of view using existing infrastructure, even with sparse coverage of roads in urban regions. We address these challenges through our two key contributions:
\vspace{-5pt}
\begin{itemize}
    \item A repeatable approach for generating realistic fine-grain synthetic datasets using traffic flow data within a microscopic traffic simulator, facilitating researchers with more realistic data. Our method takes readily available coarse-grain public traffic flow data. It generates a synthetic dataset using traffic data within a simulator that closely matches the coarse-grain distributions of the public traffic flow real-world dataset.
    \item A novel technique that can detect and localize a traffic incident without the incident being directly in the field of view of a visual sensor. Localization of the incident is achievable without knowing the precise distance between sensors. This incident detection technique is also robust to sparse sensor placement in urban regions.
\end{itemize}

We generated a synthetic dataset for Tempe, for 12 separate urban backbone roads for an area of about 4 square miles\cite{noauthor_traffic_nodate} with a traffic approximation model and confirmed by the Kolmogorov-Smirnov test\cite{kstest}. TabNet\cite{tabnet} models were trained on 31 days of simulated data. In an urban region, \inet{} detection rate of traffic incidents was 98\%, the mean time to detect incidents was 197.44 seconds, and the false alarm rate was a mere 6.26\% with a sensor sparsity of 81.4\%. Furthermore,  applied to a highway scenario, \inet{} achieved a detection rate of 99\% with a false alarm rate of 4.17\%, making it suitable to tackle both environments.
% The implementation of our work is available on our \href{https://github.com/MPSLab-ASU/IncidentNet}{GitHub} repository.

\section{Related Works}
\label{sec: related works}
% \kh{Reduce related works section to one column}
\subsection{Challenges of Macroscopic Datasets} 
The PEMS\cite{PeMSwebsite} Bay dataset collects traffic data using inductive loop detectors placed throughout the highways in the Bay area and other parts of California. Traffic metrics like average speed, occupancy, and vehicle count are gathered and aggregated at 5-10 minute intervals without distinguishing information about individual vehicles. I-880\cite{skabardonis1996880} and METR-LA\cite{metr_la} also capture macroscopic data through inductive loop detectors, similar to the PEMS dataset. These datasets (i) don't capture the nuance details essential for better accuracy detection in urban areas, and (ii) primarily originate from highway and freeway sensors, not reflecting urban-level traffic dynamics, making it difficult to build accurate incident detection systems. We address these challenges by simulating fine-grained traffic using microscopic traffic simulation built on real-world coarse datasets.

\subsection{Limitations of Existing Incident Detection and Localization Methods}
% |C{1.7cm}|C{1.5cm}|C{2cm}|C{1.5cm}||C{1.5cm}||C{1.5cm}|
\begin{table}[!htbp]
\centering
\begin{tabular}{|c|c|c|c|c|c|}
\hline
\textbf{Work} & \textbf{Region} & \textbf{Dataset}& \textbf{DR} & \textbf{FAR} & \textbf{MTTD} \\ \hline
\cite{liang_traffic_2022} & Highway & Macroscopic & 88.09 \% & 2.80 \%  & 26.80 sec \\ \hline
\cite{chen_more_2023} & Highway & Macroscopic & 99.33 \% & 6.50 \%  & NA \\ \hline
\cite{han_traffic_2020} & Urban & Microscopic & 86.40 \% & 8.69 \%  & 61 sec \\ \hline
\cite{zhu_deep_2018} & Urban & Macroscopic & 86.6 \% & 5.12 \%  & NA \\ \hline
\cite{yang_traffic_2023} & Both & Macroscopic & 80 \% & 4.68 \%  & 450 sec \\ \hline
\cite{atilgan2023towards} & Highway & Macroscopic & 74 \% & 7.6 \%  & 300 sec \\ \hline
\textbf{Ours} & \textbf{Both} & \textbf{Microscopic} & \textbf{98 \%} & \textbf{6.26 \%}  & \textbf{197.4 sec} \\ \hline
\end{tabular}
\caption{Summary of incident detection works and their observed metrics. Given our interest in urban regions, \cite{zhu_deep_2018} has shown the best detection and false alarm rates.}
\label{tab: related_works}
\vspace{-5pt}
\end{table}

Various incident detection algorithms and their metrics like Detection Rate (DR), False Alarm Rate (FAR), Mean time to detect (MTTD), region (type of road), and type of data (microscopic and macroscopic) have been summarized in Table \ref{tab: related_works}. \cite{liang_traffic_2022} used multiple highway cameras to detect incidents via spatial trajectory anomalies but did not address complex scenarios like ramps or lane closures. \cite{chen_more_2023,xu_ensemble_2024} used the XGBoost algorithm for highway incident detection with \cite{xu_ensemble_2024} also calculating incident severity. However, they make predictions every 5 minutes, introducing increased incident detection time. In urban settings, \cite{yu_arterial_2015} and \cite{han_traffic_2020} detected incidents using comparative and pattern-matching approaches with thresholds but failed to work well in dynamic traffic conditions, and they also require the installation of additional infrastructure to enable communication. Alternatively, \cite{zhu_deep_2018} utilized a deep learning approach using traffic volume data from inductive loop detectors to detect incidents. However, its reliance on an adjacency matrix representing a sensor network and using macroscopic data raises scalability and efficiency concerns. Also, similar to \cite{xu_ensemble_2024}, they predict incidents at 5-minute intervals, leading to delayed incident detection.

Incident detection algorithms reliant on data from all sensors during inference face efficacy challenges as some sensors may become non-functional over time. This was shown in the report \cite{bikowitz_evaluation_nodate}, which highlighted that about 25\% of New York's traffic sensors were nonfunctional during the survey. This has not been a focus area in previous studies, making it a crucial problem to be addressed.

\section{Microscopic Traffic Dataset Generation}
\label{sec: Fine Grain Dataset Generation}
Most real-world traffic flow information is macroscopic, but we need microscopic data to detect incidents accurately in urban environments. We can obtain microscopic data through simulators such as SUMO\cite{sumo}, VSIM\cite{yang_traffic_2023}, and AIMSUN \cite{AimsunManual}. We do this in three parts: (i) Microscopic traffic flow simulation from macroscopic data, (ii) Traffic incident simulation, and (iii) Dataset generation.

\subsection{Microscopic Traffic Flow Simulation from Macroscopic Data}
\begin{figure}[!htbp]
    \vspace{-5pt}
    \centering
    \includegraphics[scale = 0.3]{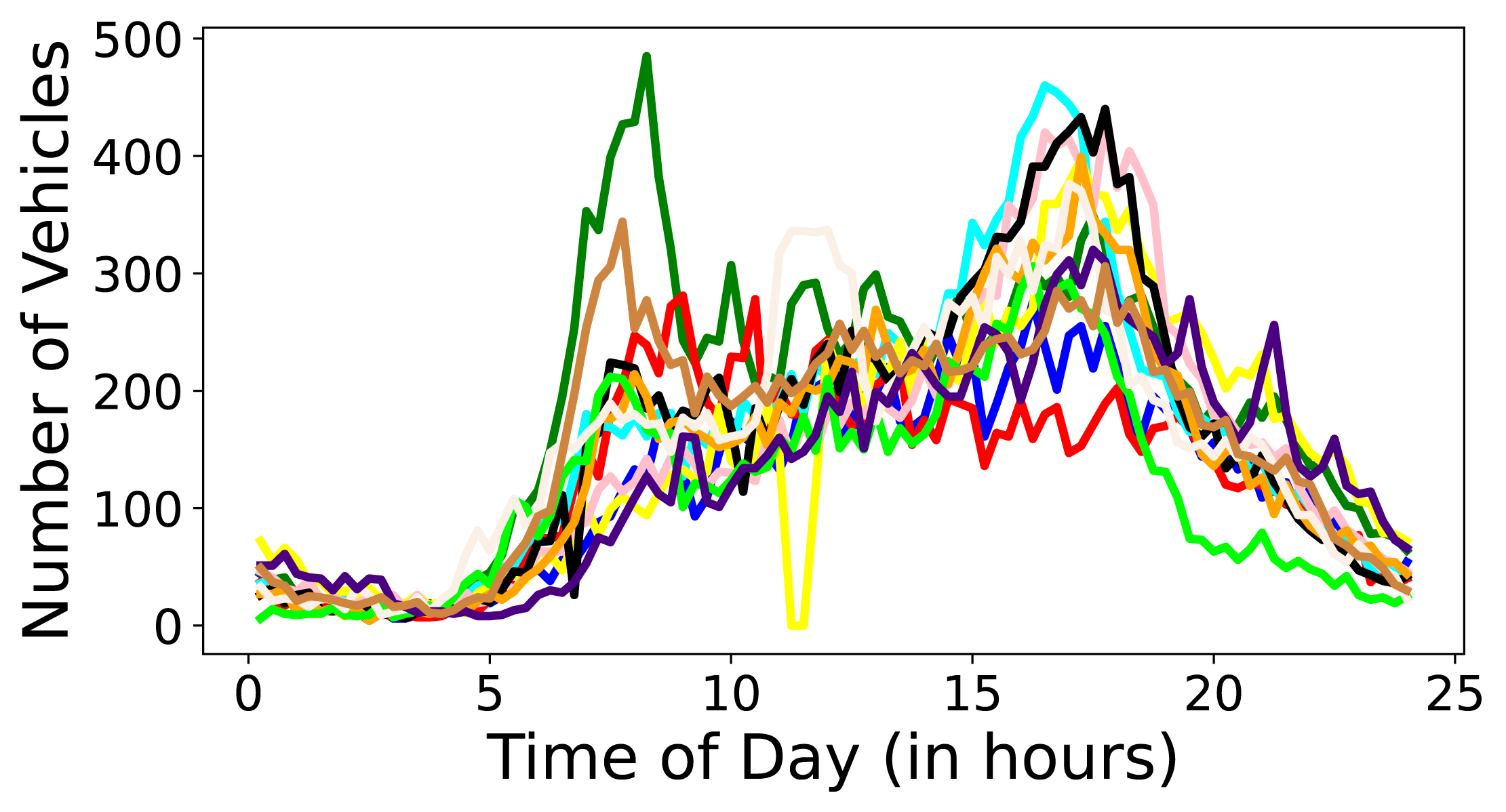}
    \caption{The plot of the vehicle counts for a 24-hour period from the Department of Transportation of Tempe for the 12 roads between the placed sensors of interest from the selected Tempe region shown in Fig. \ref{fig: tempe map with real-world sensor placement}.}
    \label{fig: tempe traffic plot}
    \vspace{-5pt}
\end{figure}

It's essential to model macroscopic data such as publicly available vehicle counts to create realistic traffic simulations, as simulators don't have this capability inherently. The city of Tempe provides vehicle count data aggregated and reported every 15 minutes for multiple days. We use a 24-hour period of data as shown in Fig. \ref{fig: tempe traffic plot} and generate microscopic traffic information that can produce vehicle counts for every second, allowing simulators to use this data to simulate the traffic. We start by computing the average vehicle counts across all roads of interest at every time step in an urban region.

\begin{figure}[!htbp]
    \centering
    \includegraphics[scale = 0.3]{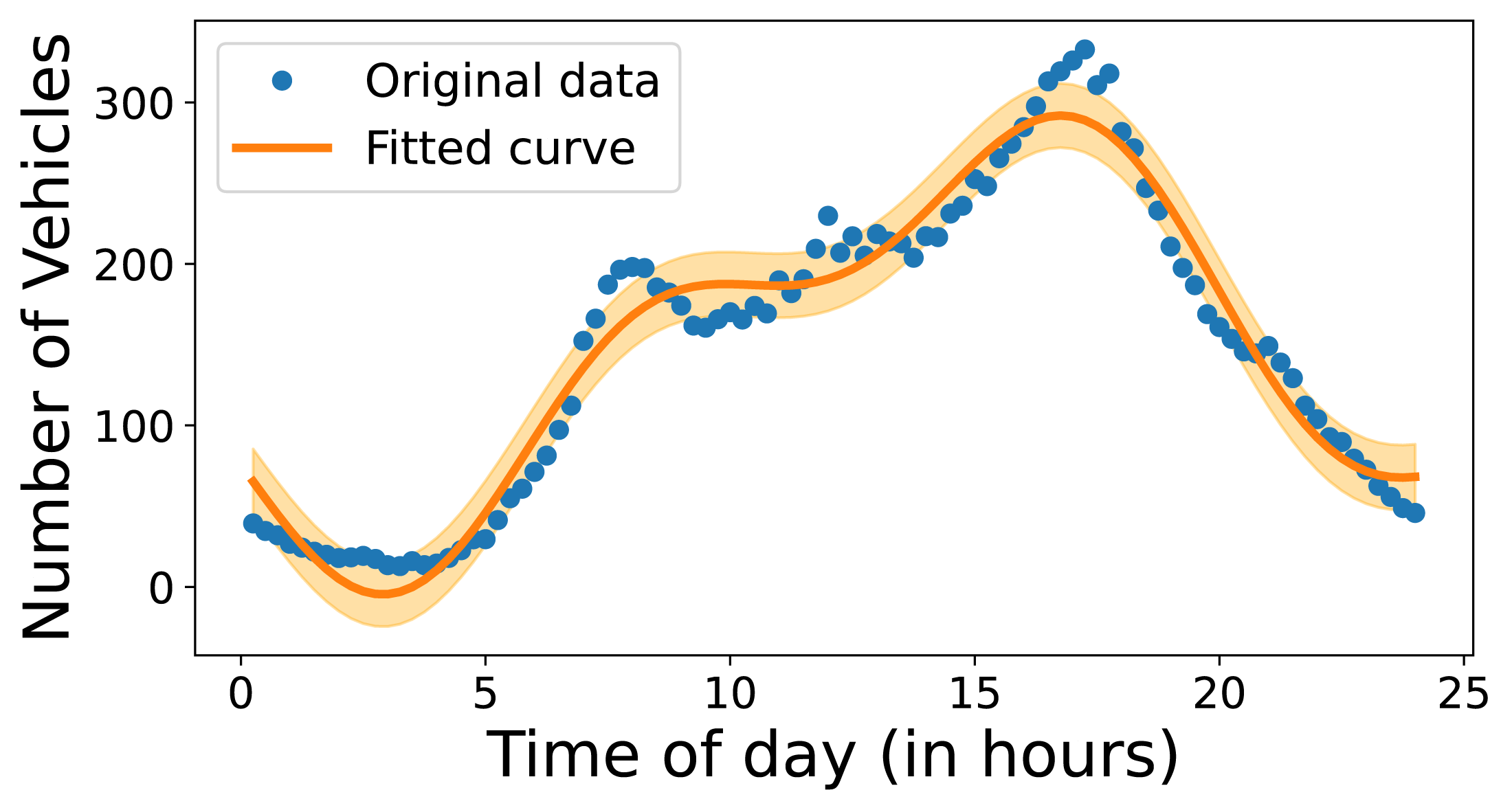}
    \caption{Representation of averaged ground-truth vehicle counts and generated traffic flow model. To ensure variance in generated vehicle counts, we add a small deviation $alpha$.}
    \label{fig: modeled traffic curve} 
    \vspace{-15pt}
\end{figure}

We then apply Fast Fourier Transforms (FFT)\cite{fft} to the averaged vehicle count data points as shown in Fig. \ref{fig: modeled traffic curve} and obtain the top two frequencies to build a non-linear equation that can approximately model the average traffic behavior over time, represented by the Equation \ref{eqn: default traffic model}.
\vspace{-2pt}
\begin{equation}
f(t) = A_1 \sin(B_1 t + C_1) + A_2 \sin(B_2 t + C_2) + D + \alpha
\label{eqn: default traffic model}
\vspace{-5pt}
\end{equation}

To determine the parameters that best represent the original vehicle counts, we use the Levenberg-Marquardt algorithm (Equation \ref{eqn: levenberg equation}) to tune the parameters, which results in minimizing the difference between the original vehicle counts and traffic flow model predictions.

\vspace{-15pt}
\begin{equation}
    \delta = \frac{J^{T}[y-f(t)]}{(J^{T}\cdot J+\lambda I)}
    \label{eqn: levenberg equation}
    \vspace{-10pt}
\end{equation}
\vspace{-5pt}

In this equation, $\lambda$ represents the damping factor ($=0.01$); $\delta$ represents the amount by which the parameters are updated in each step; $J$ is the  Jacobian matrix of partial derivative of the Equation \ref{eqn: default traffic model} with respect to its parameters; $f(t)$ represents the vehicle count that we obtain from Equation \ref{eqn: default traffic model}.
\vspace{-2pt}

\subsection{Traffic Incident Simulation}
Traffic incidents are simulated by halting vehicle(s). Depending on the likelihood of incident occurrence per vehicle, we first determine if we must insert an incident. If we have to insert an incident, we select one of the two incident types, halted vehicle and multi-vehicle crashes, for a duration also picked randomly based on the probability of the incident's severity depending on the two types of incidents. Once an incident is inserted, the radius of impact of the incident is calculated based on the severity of the incident. Inside the radius of impact, the vehicles are slowed down to emulate real-world crash behavior. It is challenging to categorize incidents as there is no direct access

\subsection{Dataset Generation}

\begin{figure}[!htbp]
    \centering
    \includegraphics[scale=0.3]{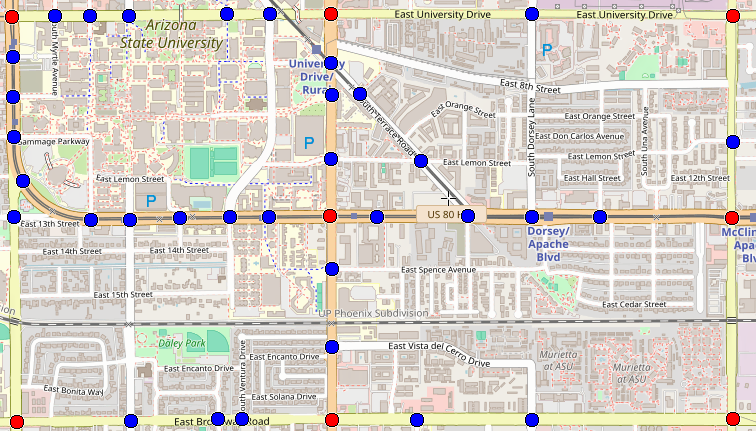}
    \caption{Shows A Tempe, AZ region selected as the test area for our implementations. All the plotted points indicate the locations where cameras can be deployed for simulation. However, the deployed locations are highlighted in red to make the deployment of cameras similar to the real world.}
    \label{fig: tempe map with real-world sensor placement}
    \vspace{-10pt}
\end{figure}

Fig. \ref{fig: tempe map with real-world sensor placement} shows all the intersections at which traffic lights and, therefore, cameras can be placed. The red dots indicate the locations where the sensors are placed to capture simulation data, leading to an inherent sparsity in data capture. The simulation process is executed for multiple days, depending on the simulation configuration. We use an API service called Traci, provided by SUMO, to extract all the available features like vehicle counts, occupancy, vehicle speed, time of the day, and vehicle identifiers within a range of sensor locations similar to cameras for every second and consolidate them into a tabular format, generating huge raw microscopic traffic flow and incident dataset.

\section{Traffic Incident Detection, Localization and Severity Estimation}
The captured raw dataset has (i) low variance as data is captured second, and traffic does not change significantly in such short intervals, leading to repeated data, (ii) frequent zero values, which are important from a data perspective but difficult to use from a deep learning perspective, like traffic counts, which makes sense for data, but acts as a sparse value for deep learning approaches and (iii) missing critical features such as vehicle travel time, limiting its effectiveness in training deep learning models. We consider data pre-processing approaches to overcome these challenges.

\subsection{Feature Extraction from Raw Data}
As travel time between intersections is an essential metric for incident detection, we used vehicle re-identification\cite{huang_machine-learning-based_2022} to compute the travel time between all possible combinations of two contiguous intersections based on the sensor placements. Incorporating these travel times, junction mean speed, vehicle count, and vehicle occupancy into our dataset resulted in a feature-rich data source, significantly improving the dataset's utility and addressing the raw dataset's challenges. 

Due to the presence of outlier data points, for example, when vehicles make unscheduled stops, we apply rolling window averages to reduce their impact. This technique involves averaging historical and current data, which allows us to smooth out anomalies in the dataset. If the current duration is labeled as an incident in the raw data, we label the rolling window average data points as incidents.

\subsection{Model Selection}
Despite these pre-processing efforts, we still observe missing data due to vehicles bypassing major intersections through interior roads and not getting re-identified. However, it still represents valuable information on traffic behavior. So, it is crucial to consider deep learning approaches that can better handle missing data.

Self-attention-based transformer models have worked exceptionally well to understand long-range sequences. TabNet\cite{tabnet} is an architecture designed for interpretable learning from tabular data. For training, the data is processed by the TabNet encoder, which uses a decision-making decoder to classify the results. Each TabNet encoder block comprises an attentive transformer block, a learnable mask, and a feature transformer. The learnable mask performs a soft selection of salient features, which are processed by the feature transformer, and the attentive transformer learns the importance of each feature during training. Multiple layers of these encoder blocks form the TabNet Encoder. The authors claim that this instance-wise feature engineering and learning allows for a better performance than Decision Tree-based models like XGBoost, making it a significant factor for us to consider this as our model architecture.
% \vspace{-5pt}
\subsection{\inet{}'s Model Architecture Design}
\begin{figure}[!htbp]
    \centering
    \includegraphics[scale=0.35]{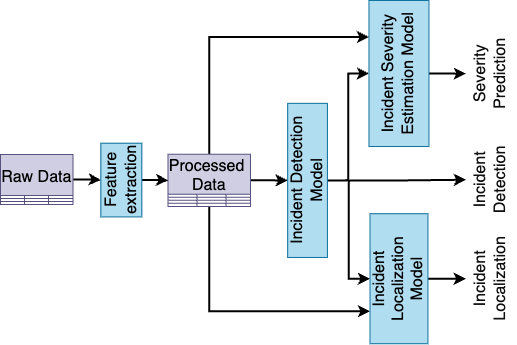}
    \caption{The block diagram depicts \inet{}'s architecture. The raw data from the simulator is transformed into processed data. For training, all data points are used for the incident detection model, and data points with positive incident labels are used for incident localization and severity estimation models. During the prediction phase, localization and severity estimation models depend on the incident detection model's prediction.}
    \label{fig: block diagram}
    \vspace{-5pt}
\end{figure}

Our incident detection architecture employs a stacked ensemble of three models dedicated to incident detection, localization, and severity estimation tasks. All three models are trained individually, with varying input data. For the incident detection model, the complete pre-processed dataset is provided as input and trained to predict if an incident has occurred in the complete selected urban region. We cannot predict the incident's class as they are not occurring in the sensor's field of view. For localization and severity estimation models, the data points with positive ground truth incident labels are considered for training. The localization model predicts the roads on which the incident occurred, and the severity estimation classifies if an incident is severe. Our ensemble model can localize and estimate severity only due to the microscopic dataset we generated.

A unique aspect of our architecture is its robustness in accommodating sparse sensor settings, a common challenge in real-world traffic monitoring scenarios. Unlike existing incident detection methods, our models are evaluated under various levels of sensor sparsity to assess the performance of each task under various degrees of sparsity.

\section{Experimental Setup}
\label{sec: Experimental Setup}
\subsection{Simulation Setup for Dataset Generation}
We generate simulation files using the OSM Web Wizard for a continuous period of 30 days to simulate traffic flow for the selected Tempe region and generate the microscopic data using the process described in our approach.

\subsection{Pre-processing Raw Dataset}
We test with three variations: 300, 600, and 900 seconds to pre-process the raw data and select the rolling window size. We train our model using pre-processed data aggregated using different window sizes and observe F1 scores of 93.12\%, 96\%, and 96\%, respectively. Given that more data increases the computational requirement, we choose 600 seconds as our window size choice, as the F1 score for 600 seconds and 900 seconds are similar.

\subsection{Model Training and Evaluation Considerations}
\label{subsec: model training and evaluation considerations}
We used TabNet to evaluate the Tempe dataset. The model was trained on NVIDIA RTX 5000 GPU, and for TabNet, the hyperparameters used are mentioned in Table \ref{tbl: tabnet hyperparams}.

\begin{table}[!htbp]
\centering
\begin{tabular}{| C{4.3cm} | C{3cm} |}
\hline
\textbf{Hyper-parameters} & \textbf{Value} \\ \hline
Prediction Layer Dimension & 64 \\ \hline
Attention Embedding Dimension & 64 \\ \hline
Optimizer Momentum  & 0.3 \\ \hline
Optimizer & Adam \\ \hline
Learning Rate & 0.02 \\ \hline
Epochs & 80 \\ \hline
Loss Function & Cross Entropy \\ \hline
\end{tabular}
\caption{Model training hyper-parameters for TabNet.}
\label{tbl: tabnet hyperparams}
\vspace{-5pt}
\end{table}

Table \ref{tbl: traffic_metrics} shows the different metrics we use to evaluate the performance of our model. We used the three standard metrics, Detection Rate (DR), Mean Time to Detect (MTTD), and False Alarm Rate (FAR), to evaluate the performance of the incident detection algorithm.

\begin{table}[!htbp]
\centering
\renewcommand{\arraystretch}{1.5}
\begin{tabular}{|c|c|}
\hline
\textbf{Metrics} & \textbf{Definition} \\ \hline
DR (Detection Rate) & $\frac{TP}{TP + FN}$ \\ \hline
FAR (False Alarm Rate) & $\frac{FP}{FP + TN}$ \\ \hline
Accuracy  & $\frac{TP + TN}{TP + TN + FP + FN}$ \\ \hline
Precision & $\frac{TP}{TP + FP}$ \\ \hline
Recall & $\frac{TP}{TP + FN}$ \\ \hline
F1 Score & $2 \times \frac{\text{Precision} \times \text{Recall}}{\text{Precision} + \text{Recall}}$ \\ \hline
Specificity & $\frac{TN}{TN + FP}$ \\ \hline
\end{tabular}
\caption{Traffic Incident Detection Metrics and their definitions based on confusion matrix, where TP = true positives, TN = true negatives, FP = false positives, FN = false negatives.}
\label{tbl: traffic_metrics}
\end{table}
\vspace{-5pt}

\section{Results}
\subsection{Our Microscopic Data Matches Very Well with Real-World Macroscopic Data}
\label{sec: Validation of Synthetic Data}

To validate that our simulated data accurately reflects real-world conditions in the Tempe region, we aggregated the microscopic simulation data to match the time frame of Tempe's macroscopic real-world traffic count data. This produces a distribution similar to the original data represented in Fig. \ref{fig: modeled traffic curve}. To assess the similarity, we used the Kolmogorov-Smirnov (KS) test\cite{kstest}, which evaluates the similarity between two distributions by calculating two metrics: KS statistic and p-value. The KS statistic measures the maximum discrepancy between the distribution functions of datasets. The p-value measures the probability of low discrepancy between the two datasets. The null hypothesis is true when both distributions are similar. We reject the null hypothesis if the p-value is below the accepted significance of 0.05.

\begin{figure}[!htbp]
    \centering
    \includegraphics[scale = 0.3]{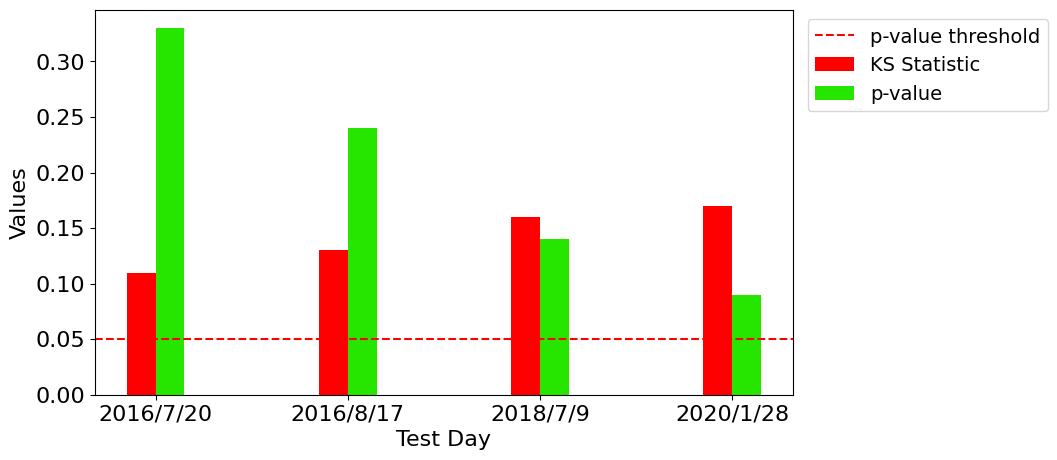}
    \caption{The KS Statistic and the p-value obtained from the KS test for the four days of data made available by Tempe are shown. The p-value threshold is indicated as the red line.}
    \label{fig: ks_test.png}
    \vspace{-10pt}
\end{figure}

The Tempe Department of Transportation provides the vehicle count data for just four days, and the days on which they were collected are randomly presented. Of the 30-day simulated data, we selected four days randomly for validation. We observed that, though there is variation in the KS Statistic and the p-value, all of them pass the cut-off according to the algorithm as shown in Fig. \ref{fig: ks_test.png}, indicating similarity between the simulated and original data.

\subsection{\inet{} is Better at Detecting Incidents in Urban Regions Compared to the Previous Works}

\begin{table*}[!htbp]
\vspace{5pt}
\centering
\begin{tabular}{|c|c|c|c|c|c|c|c|c|c|}
\hline
\textbf{Algorithm} & \textbf{DR} & \textbf{MTTD} & \textbf{FAR} & \textbf{Accuracy} & \textbf{Precision} & \textbf{Recall} & \textbf{F1 Score} & \textbf{AUC-ROC} & \textbf{Specificity} \\ \hline
Our approach (XGBoost) & 96 \% & 94 secs & 11.03 \% & 92.13 \% & 93.76 \% & 90.49 \% & 87.43 \% & 91.46 \% & 95.7 \% \\ \hline
Our Approach (TabNet) & 98 \% & 197.44 secs & 6.26 \% & 93.85 \% & 94.27 \% & 91.17 \% & 92.70 \% & 93.51 \% & 95.95 \% \\ \hline
Zhu et al. \cite{zhu_deep_2018} & 51 \% & 471 secs & 35.42 \% & 60.06 \% & 40 \% & 50.45 \% & 44.62 \% & 51 \% & 64.57 \% \\ \hline
\end{tabular}
\caption{The table compares the previous state-of-the-art, XGBoost and our approach for the microscopic dataset generated for urban traffic scenarios. Our approach performed exceptionally well when compared to the previous state-of-the-art. The other outcome we observed was that XGBoost performed better than the state-of-the-art, proving the importance of microscopic datasets. Our method predicted incidents every 30 seconds instead of every 5-minute interval, as in \cite{zhu_deep_2018}.}
\label{tab:modeleval}
\vspace{-15pt}
\end{table*}

As highlighted before, a fast and accurate traffic incident detection algorithm can reduce the impact of incidents economically and environmentally and, mainly, reduce fatality rates. Their impact can be evaluated using metrics such as DR, FAR, and MTTD, which are defined in \ref{subsec: model training and evaluation considerations}. We evaluated our work against the state-of-the-art by training a model using the architecture provided by \cite{zhu_deep_2018}, which we implemented to the best of our understanding as the official model implementation was not available, and the XGBoost model architecture, using our microscopic dataset. As XGBoost has proven to work well on tabular data\cite{chen_more_2023,xu_ensemble_2024} due to its efficient selection of global features with high information value \cite{decisiontrees}, to assess the impact of the microscopic dataset, we also evaluate with XGBoost as the model consideration in our approach.

We evaluated all the models on a newly generated evaluation dataset for the same region, consolidated in Table \ref{tab:modeleval}. We observe that XGBoost's performance improves drastically compared to the model's performance on microscopic data, showing the importance of considering microscopic datasets for traffic incident detection. Our TabNet approach is more accurate than XGBoost, with a DR of 98\% and FAR of 6.26\%. The downside we observed is that the MTTD is 197.44s, almost 100s higher than XGBoost. However, this is offset by the much lower FAR, which indicates that our model has the ability to report incidents more accurately while remaining fast enough to be within the 7-minute mark, as defined in \cite{emergency_response}. The inference time of the TabNet model on the Intel Xeon W-2555 CPU was 5 ms, providing timely insights and supporting real-time decision-making.

\subsection{\inet{} Works Even In Sparse Sensing Condition}
\begin{figure}[!htbp]
    \centering
    \includegraphics[scale = 0.3]{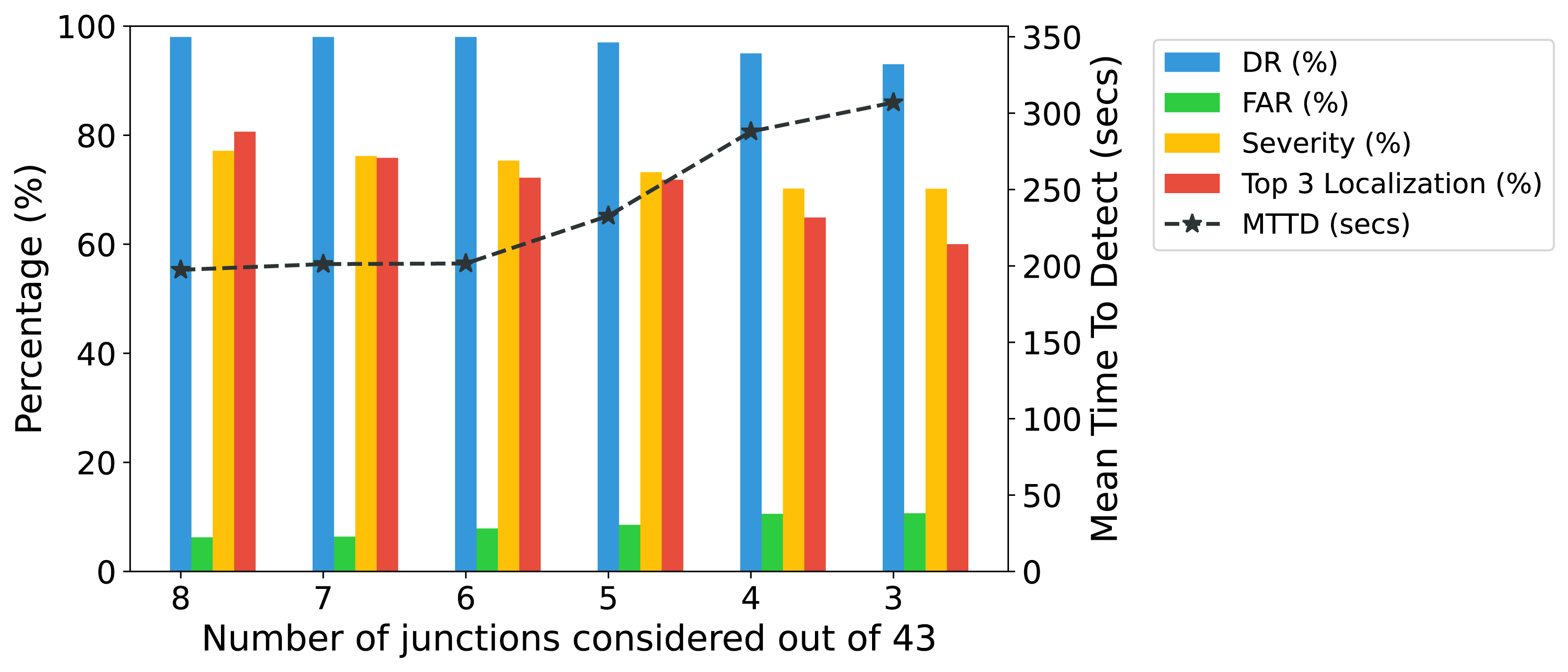}
    \caption{Our approaches' performance with consideration for different sparsity levels. Notice that the incident detection rate is still high for sparsity, as high as 93\%. Tabnet performs better in incident detection with a low false alarm rate.}
    \label{fig: traffic_incident_detection_metrics}
    \vspace{-8pt}
\end{figure}

Sensing hardware is fallible and degrades over time, thus it is reasonable to assume that not all cameras will be working at all times. It is vital for a model to be able to work even in such conditions. So, we test our model's performance with increasing levels of sparsity. We start with a realistic sensor deployment at 8 of the 43 possible intersections and scale down to just 3 intersections.

In Fig. \ref{fig: traffic_incident_detection_metrics}, we observe with increased sensor sparsity that our model still retains the capability to detect if an incident occurs, but the accuracy of localization and severity predictions is reduced. Interestingly, with only six sensors, the MTTD does not increase much. However, the MTTD increases more drastically with fewer than 6 sensors. The FAR also increases with an increase in sparsity. Although we observe this increase, we show our model is still capable of predicting metrics, even during infrastructure anomalies.

\subsection{\inet{} can Detect Incidents on Highways}

\begin{table}[!htbp]
\vspace{5pt}
\centering
\begin{tabular}{|c|c|c|c|}
\hline
\textbf{Algorithm} & \textbf{DR (\%)} & \textbf{FAR (\%)} & \textbf{MTTD (secs)} \\ \hline
Our Approach (XGBoost) & 98 & 6.02 & 45   \\ \hline
Our Approach (TabNet)  & 99 & 4.17 & 70   \\ \hline
\end{tabular}
\caption{Highway performance of \inet{} our approach compared against the XGBoost model on our microscopic dataset. Results demonstrate that the performance of XGBoost improves because of the microscopic dataset, and \inet{} performs better than XGBoost.}
\label{tab:highway}
\vspace{-15pt}
\end{table}

Given that our model works in urban regions, we test if our approach works in a highway scenario. We used an 8-mile highway stretch, inserted the sensors on every available ramp, and simulated the microscopic dataset. We trained and evaluated using XGBoost and our model. The metrics obtained are shown in Table \ref{tab:highway}. XGBoost model performed better than the previous works shown in Table \ref{tab: related_works}. Our model performed better than XGBoost in terms of DR and FAR, with a very minimal increase in the MTTD, proving that our approach works in both urban regions and highways.

\section{Conclusion}
In this paper, we have shown that \inet{} can successfully detect traffic incidents with a high detection rate in urban roads using microscopic sensor data. In particular, the results confirm that using just 3 instrumented intersections of the 43 possible \inet{} can accurately detect, localize, and classify incidents in a large area, marking a significant advancement in traffic management technologies. Building upon this supervised model, a promising next step is implementing a semi-supervised version of \inet{}. This would allow the model to continually improve and handle recurring congestion when deployed in real-world settings. This work also highlights the importance of sensor placement in sparse sensing scenarios, highlighting the need for an algorithm to efficiently place sensors while maximizing the incident detection rate in sparse sensing. Further investigation could extend to categorizing incidents into more classes and enhancing localization accuracy, possibly including rough estimation of distances of incidents from the intersections. Additionally, extending to other regions is part of the future scope of this work.

% {
% \bibliographystyle{plain}
% \bibliography{root.bbl}
% }
% \section*{References}
\bibliographystyle{plain} % Keep the same style

 % Include the .bbl file directly

\end{document}